\documentclass[11pt]{article}
\usepackage[]{graphicx}
\usepackage[]{color}

\pdfoutput=1

\makeatletter
\def\maxwidth{ %
  \ifdim\Gin@nat@width>\linewidth
    \linewidth
  \else
    \Gin@nat@width
  \fi
}
\makeatother

\definecolor{fgcolor}{rgb}{0.345, 0.345, 0.345}

\usepackage{framed}
\makeatletter
 {\par\unskip\endMakeFramed%
 \at@end@of@kframe}
\makeatother

\definecolor{shadecolor}{rgb}{.97, .97, .97}
\definecolor{messagecolor}{rgb}{0, 0, 0}
\definecolor{warningcolor}{rgb}{1, 0, 1}
\definecolor{errorcolor}{rgb}{1, 0, 0}
\newenvironment{knitrout}{}{} 

\usepackage{alltt} 
\usepackage{amsmath}
\usepackage{graphicx}
\usepackage{color}
\usepackage[round]{natbib}
\usepackage[linesnumbered]{algorithm2e}
\usepackage[
         colorlinks=true,
         linkcolor=black,
         citecolor=black,
         urlcolor=black]
         {hyperref}

\definecolor{darkgreen}{rgb}{0,0.6,0}
\definecolor{darkred}{rgb}{0.6,0.0,0}
\definecolor{lightbrown}{rgb}{1,0.9,0.8}
\definecolor{brown}{rgb}{0.6,0.3,0.3}
\definecolor{darkblue}{rgb}{0,0,0.8}
\definecolor{darkmagenta}{rgb}{0.5,0,0.5}

\newcommand{\pkg}[1]{\textsf{#1}}

\IfFileExists{upquote.sty}{\usepackage{upquote}}{}

\begin{document}

\title{Futility Analysis in the Cross--Validation of Machine Learning Models}
\author{Max Kuhn ({\tt  max.kuhn@pfizer.com}) \\
        Nonclinical Statistics\\
       Pfizer Global R$\&$D\\
       Groton, CT 06340, USA}

\maketitle

\begin{abstract}
Many machine learning models have important structural tuning parameters that cannot be directly estimated from the data. The common tactic for setting these parameters is to use resampling methods, such as cross--validation or the bootstrap, to evaluate a candidate set of values and choose the best based on some pre--defined criterion. Unfortunately, this process can be time consuming. However, the model tuning process can be streamlined by adaptively resampling candidate values so that settings that are clearly sub-optimal can be discarded. The notion of {\em futility analysis} is introduced in this context. An example is shown that illustrates how adaptive resampling can be used to reduce training time. Simulation studies are used to understand how the potential speed--up is affected by parallel processing techniques. 
\end{abstract}

\vspace{.1in}


\vspace{.1in}

\section{Introduction}\label{S:intro}

Machine learning \citep{Bishop2007} uses past data to make accurate predictions of future events or samples. In comparison to traditional inferential statistical techniques, machine learning models tend to be more complex black--box models that are created to maximize predictive accuracy \citep{Breiman:2001wl}. The process to develop these models is usually very data--driven and focuses on performance statistics calculated from external data sources. 

The effectiveness of a model is commonly measured using a single statistic which we will refer to here as {\em model fitness}. For models predicting a numeric outcome, the fitness statistic might be the root mean squared error (RMSE) or the coefficient of determination ($R^2$). For classification, where a categorical outcome is being predicted, the error rate might be an appropriate measure. 

When creating machine learning models, there are often {\em tuning parameters} or {\em hyper parameters} to optimize. These are typically associated with structural components of a model that cannot be directly estimated from the data.  For example:
\begin{itemize}
\item A $K$--nearest neighbor classifies new samples by first finding the $K$ closest samples in the training set and determines the predicted value based on the known outcomes of the nearest neighbors. In this model, $K$ (and possibly the distance metric) are tuning parameters.
\item In classification and regression trees, the depth of the tree must be determined. Many tree--based models first grow a tree to the maximum size then prune tree back to avoid over--fitting. \citet{cart} use cost--complexity pruning for this purpose and parameterize this setting in terms of the complexity value $c_p$. \citet{quinlan1993c4} uses an alternative pruning strategy based on uncertainty estimates of the error rate. In this case, the confidence factor (CF) is a tuning parameter that determines the depth of the tree.
\item Partial least squares models \citep{wold2001pls} utilize the data in terms of latent variables called PLS components. The number of components must be determined before the final PLS model can be used. 
\end{itemize}
In many cases, the values of the tuning parameters can have a profound effect on model efficacy. \citet{Delgado14} describe a study where a large number of classifiers were evaluated over a wide variety of machine learning benchmark data sets. Their findings validate the importance of model tuning. Despite their importance, it is rare that reasonable values of these parameters are known {\it a priori}.

To determine appropriate values of the tuning parameters, one approach is to use some form of resampling to estimate how well the model performs on the training set \citep{apm}. Cross--validation, the bootstrap  or variations of these are commonly used for this purpose. A single iteration of resampling involves determining a subset of training set points that are used to fit the model and a separate ``holdout'' set of samples to estimate model fitness. This process is repeated many times and the performance estimates from each holdout set are averaged into a final overall estimate of model efficacy. 

There are different types of resampling methods, the most common types being: $k$--fold cross--validation, repeated $k$--fold cross--validation, leave--one--out cross--validation, Monte Carlo cross--validation and the bootstrap \citep{apm}. Each resampling scheme has its own variance and bias properties. For example, the bootstrap tends to have small variance but substantial bias while traditional $k$--fold cross--validation has small bias but high uncertainty (depending on $k$). Recent research \citep{Molinaro:2005p47,Kim:2009im} suggests that repeating $k$--fold cross--validation is advisable based on having acceptable variance and bias in comparison to the other approaches.  

Denote the training data as $D$. If there are $B$ resamples, denote the resampled version of the data as $R_{i}$ and the holdout set induced by resampling as $T_{i}$, where $i = 1\ldots B$. The complete set of tuning parameters is symbolized by $\Theta$ and an individual candidate set of tuning parameters by $\theta_j$ with $j=1\ldots p$. Some models have multiple tuning parameters and, in these cases, $\theta$ is vector valued. From each resample and candidate parameter set, let the fitted model be $\hat{f}_{ij}(R_{i};\theta_j)$ and the resulting estimate of model fitness be denoted as $Q_{ij}$. The  resampled estimate of fitness for each parameter set is $\hat{Q}_j = 1/B\sum_{i=1}^B Q_{ij}$. 

The grid search strategy outlined in Algorithm \ref{A:Resamp} is one possible approach for optimizing the tuning parameters. First, the set of candidate  values  $\Theta$  is determined along with the type of resampling and the number of data splits. For each parameter combination, the model fitness is estimated via resampling and the relationship between the tuning parameters and model performance is characterized. From this, a rule for choosing $\theta_{opt}$,  based on the resampling profile, is needed. The choice can be made based on the empirically best result or by some other process, such as the one--standard error rule of \citet{cart}. After choosing $\theta_{opt}$, one final model $\hat{f}(D;\theta_{opt})$ is created using the optimized settings and the entire training set $D$. Grid search is not the only approach that can be used to optimize tuning parameters. For example, \citet{Ustun:2005ew} used evolutionary search procedures to optimize the performance of a support vector machine regression model. 

\begin{algorithm}
\label{A:Resamp}
\caption{The nominal grid search process for tuning a model using resampling.}
\SetAlFnt{\tiny\sf}
Define parameter set $\Theta$\;
\For{$i =1 \ldots B$}{\nllabel{test}
    Generate $R_{i}$ and $T_{i}$\;
    \For{$j=1\ldots p$}{
         Fit $\hat{f}_{ij}(R_{i}; \theta_p)$\;
         Predict $T_{i}$ to estimate $Q_{ij}$\;
    }    
}
Calculate $\hat{Q}_1\ldots \hat{Q}_p$\;
Determine $\theta_{opt}$\;
Fit the final model $\hat{f}(D; \theta_{opt})$\; 
\end{algorithm}

\citet{krstajic2014cross} discuss resampling in the context of model tuning and describe potential pitfalls.  They review historical publications and also differentiate between {\it cross--validatory choice} and {\it cross--validatory assessment}. The goal of the former is to choose between sub--models (e.g. a 3--nearest neighbor versus 5--nearest neighbor model). The latter is focused on an accurate assessment of a single model. In some cases,  both are important and this manuscript focuses on choosing a model and estimating performance to an acceptable level of precision.

To illustrate the process of model tuning, a data set for predicting whether a chemical compound will damage an organism's genetic material, otherwise known as {\em mutagenicity}, was used \citep{Kazius:2005up}. They labeled 4335 compounds as either a mutagen or non--mutagen. We generated 830 descriptors of molecular structure \citep{Leach:2007tq} for each compound and used these as predictors of mutagenicity. Examples of the descriptors used in these analyses are atom counts, molecular weight, surface area and other measures of size and charge. Using a predictive model, future compounds can be assessed for their potential toxicity based on these properties.

A support vector machine (SVM) classification model with a radial basis function kernel  \citep{vapnik2010nature} is used to illustrate parameter tuning. There are two tuning parameters: the radial basis function scale parameter $\sigma$ and the cost value associated with the support vectors. However, \citet{Caputo} describe an analytical formula to estimate $\sigma$ from the training set and this method was used to eliminate $\sigma$ from the tuning grid. As a result, $\Theta$ is one dimensional and the candidate set of cost values  consisted of 21 settings ranging from  0.25 to 256 on the log$_2$ scale, i.e. $\Theta = \left\{2^{-2}, 2^{-1.5} \ldots, 2^{8}\right\}$.

\begin{figure}[t]
  \begin{center}  
\begin{knitrout}
\definecolor{shadecolor}{rgb}{0.969, 0.969, 0.969}\color{fgcolor}
\includegraphics[width=.8\textwidth]{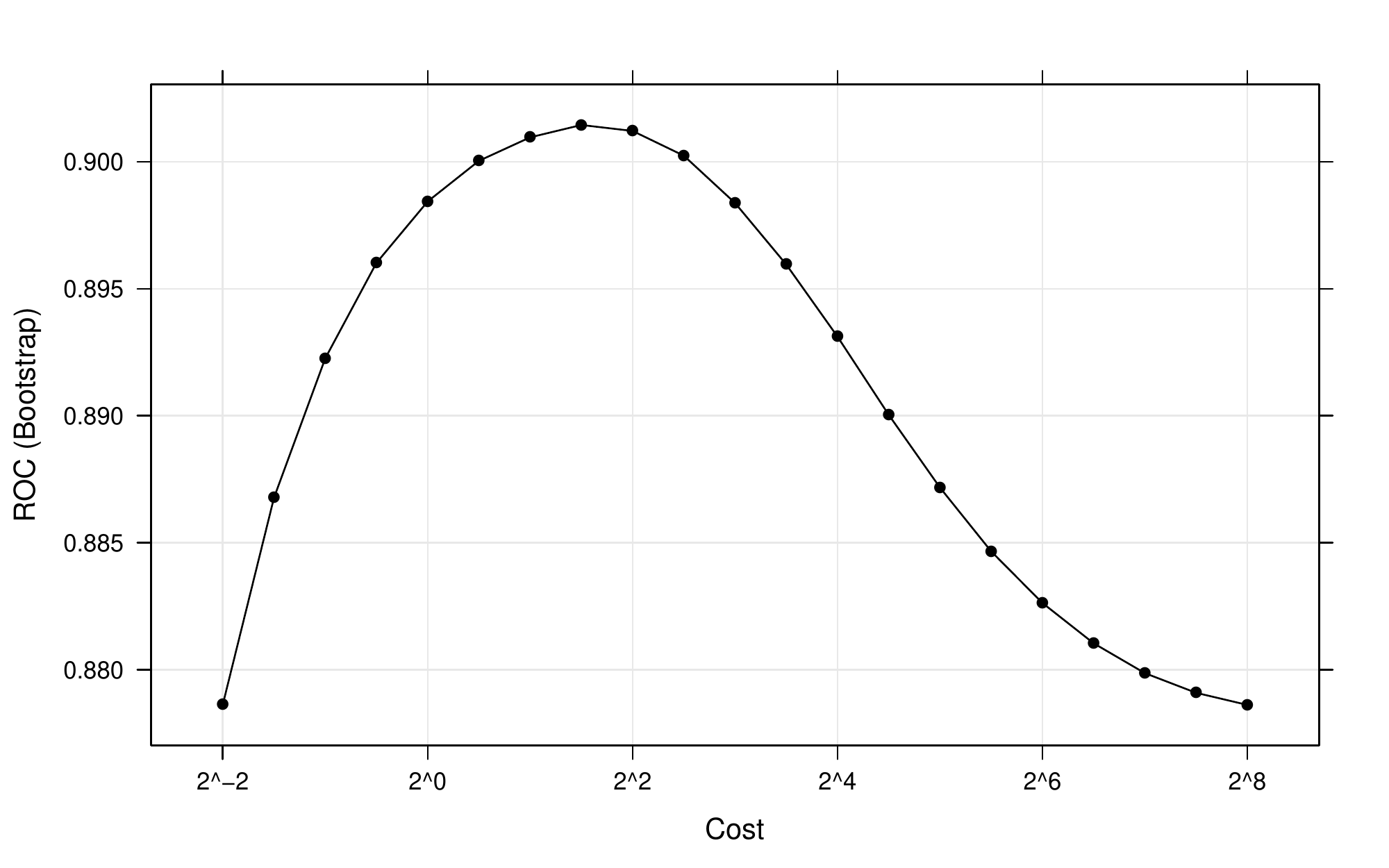} 

\end{knitrout}

    \caption{The relationship between the SVM tuning parameter and the cross--validated estimate of the area under the ROC curve. Each point is the average of 50 estimates from the resampled ROC curves.  Using a ``pick--the--winner'' strategy, $\theta_{opt}=2^{1.5}$.}
    \label{F:qsar_svm_cv}
  \end{center}
\end{figure}

Simple bootstrap resampling \citep{Efron:1983ul} was used to tune the model where, on average, the number of samples held out at each iteration of resampling was 1598. To evaluate how well the model performed within each resampling iteration, an ROC curve \citep{Altman:1994uv,Fawcett:2006gr,Brown:2006wp} is created by applying $\hat{f}_{ij}(R_i, \theta_j)$ to $T_i$. The area under the ROC curve is then used to quantify model fitness.  The results of this process are shown in Figure \ref{F:qsar_svm_cv}. When the cost value is small, the model has poor performance due to under--fitting. After a peak in performance is reached,  the model begins to become too complex and over--fit. A simple ``pick--the--winner'' strategy would select a model with a cost value of $2^{1.5}$. This sub--model is associated with an area under the curve of 0.901. With the complexity of the model now determined, the final SVM model is created with this value, the estimate of $\sigma$ and the entire training set.

One issue with this approach is that all model parameters are treated with equal priority  even after substantial evidence is available. For example, for these data, it is very clear that $\theta=2^{8}$ is highly unlikely to yield optimal results. Despite this, all $B\times p$ models are created before the the relationship between the parameters and the outcome are considered. This leads to excessive computations and, for large data sets and/or computationally expensive models, this can drastically slow down the tuning process. Additionally, some feature selection techniques are {\em wrappers} around machine learning models \citep{Kohavi:1997wca} that search for small combinations of predictors that optimize performance. If a model has important tuning parameters, this resampling process might occur at each step in the subset selection search. In this situation, increasing the computational efficiency can have a major impact on the overall time to develop a model.  

The remainder of the manuscript outlines an adaptive resampling scheme that can be used to find acceptable values of the tuning parameters with fewer model fits. The mutagenicity data are further analyzed with the proposed methodology. Simulation studies are also used to characterize the efficacy and efficiency of these approaches. Finally, the effect of parallel computations on adaptive resampling techniques are studied.

\section{Adaptive Resampling via Futility Analysis}

As resampling progresses, there may be some parameter values that are unlikely to be chosen as the optimal settings. Our goal is to identify these as early as possible so that unneeded computations can be avoided. The situation is somewhat similar to futility analysis in clinical trials \citep{Lachin:2005uu}. Whereas futility in clinical investigations is defined as ``the inability of the trial to achieve its objectives'' \citep{Snapinn:2006ih}, we might consider a tuning parameter as futile if it is unlikely to have optimal performance. 

However, there are several differences between classical futility analysis and parameter tuning. Most clinical trials involve a small number of pre--planned comparisons and the trial is usually designed to have substantial power to detect pre--specified effect sizes. In our case, there may be a large number of parameter settings and thus many comparisons. Also, prior to model building, there may be little {\it a priori} knowledge of how well the model may perform. For this reason, the understanding of a meaningful difference in the model fitness values may not be known. Given these differences, it is unlikely that existing statistical techniques for clinical futility analysis can be employed for this particular problem. However, the general concept of futility is still applicable.   

Algorithm \ref{A:adaptive} outlines how a futility assessment can be incorporated into the model tuning process. The nominal tuning process is used for the first $B_{min}$ iterations of resampling. At $B_{min}$, the fitness values $Q_{ij}$ are used to assess futility for each tuning parameter. Based on these results, a pre--defined rule is used to determine which values in $\Theta$ are unlikely to be the best and these parameter settings are removed from further consideration (and no longer resampled). The number of tuning parameter settings at each stage of resampling is denoted as $p_i$. If $p_i > 1$, the resampling process continues until either $p_i = 1$ or the maximum number of resamples is reached ($i = B$). In the former case, the resampling process would continue only for $\theta_{opt}$ to measure the fitness value to the maximum precision. In the latter cases, the nominal selection process is used to determine $\theta_{opt}$ from the parameters  still under consideration (i.e. line 11 in Algorithm \ref{A:Resamp}). 

\begin{algorithm}
\label{A:adaptive}
\caption{Tuning a model using adaptive resampling.}
\SetAlFnt{\tiny\sf}
Define parameter set $\Theta$\;
\For{$i =1 \ldots B$}{
     Generate $R_{i}$ and $T_{i}$\;
     \For{$j=1\ldots p_i$}{   
          Fit $\hat{f}_{ij}(R_{i}; \theta_{p_i})$\;  
          Predict $T_{i}$ to estimate $Q_{ij}$\;
      }    
      \If{$i > B_{min}$ and $p_i > 1$}{
          Calculate $\hat{Q}_1\ldots \hat{Q}_{p_i}$\;
          Conduct futility analysis\;
          Remove parameter settings\;
          Update $p_i$\;
          \If{$p_i = 1$} {
             set $\theta_{opt}$\;
             break\;
             }
      }
    }
\lIf{$p_i > 1$}{Determine $\theta_{opt}$}\;
Fit the final model $\hat{f}(D; \theta_{opt})$\;
  \end{algorithm}

The pivotal detail in Algorithm \ref{A:adaptive} is the method for estimating futility. In the next sub--sections, two approaches are considered in detail. One technique uses the fitness values $Q_{ij}$ to measure futility while the other uses dichotomized ``scores''. In these descriptions we assume, without loss of generality, that larger values of the fitness statistic are better. 

\subsection{Measuring Futility via Linear Models}\label{S:HT}

\citet{Shen:2011vi} describe a method for assessing futility during resampling. At resampling iteration $i$, they contrast sub--models by treating the performance estimates resulting from each split as a blocked experiment and fit a linear model
\[
Q_{kj} = \mu + \tau_j + \beta_k + \epsilon_{kj} 
\]
where $\mu$ is the grand mean, $\beta_k$ is the effect of the $k^{th}$ resample ($k = 1\ldots i$), $\tau_j$ is the effect of the sub--model ($j = 1\ldots p_i$) and $\epsilon_{kj}$ are the errors, assumed to be iid $N(0, \sigma^2)$. The interest is in comparing sub--models via statistical hypothesis tests on the $\tau_j$. 

There is a strong likelihood that the resampled fitness values have appreciable {\em within--resample} correlations where fitness values resulting from one data split tend to have a higher correlation with one another when compared to fitness values generated using a different split of the data. If this factor is not taken into account, it is likely that any inferential statements made about different parameter values may be underpowered or inaccurate. Rather than estimating the within--resample correlation, the block parameter in their model is used to account for the resample--to--resample effect in the data. 

\citet{Shen:2011vi} focused on testing $H_0: \tau_j = \tau_{j'}$ versus $H_1: \tau_j \ne \tau_{j'}$ for all $j \ne j'$ and suggests removing all ``dominated models'' from further evaluation. They also used multiple comparison corrections to account for repeated testing. The confidence level $\alpha$ is a tuning parameter for Algorithm \ref{A:adaptive} and can be used to control the greediness of the adaptive procedure.

In this manuscript, a variation of this approach is proposed. First, we suggest using one--sided hypotheses  where the current best setting is determined and the other sub--models are compared to this setting. This should improve the power of the comparisons. Secondly, our approach does not attempt to compensate for multiple testing. The confidence level controls the aggressiveness that the algorithm will eliminate sub--models. The family--wise error rate guards against any false positive findings which, in this context, has limited relevance. Additionally, \citet{Shen:2011vi} cast doubt that, after correction, the nominal significance level under the null hypothesis is really $\alpha$ when being used in this manner. Finally, instead of blocking on the splits, our model directly estimates the within--resample correlation. Specifically, at iteration $k$, we model
\begin{equation}\label{E:delta}
Q_{kj} = \mu + \tau_j + \epsilon_{kj} 
\end{equation}
where the errors are assumed to have a normal distribution with mean zero and a block diagonal covariance matrix $\Sigma$. The blocks are of size $p_i \times p_i$ and defined as $\Sigma_k = \sigma^2 (1-\rho) I_{p_i} + \sigma^2\rho J_{p_i}$. This is an exchangeable (or compound--symmetric) covariance structure where 
\[
Cov[\epsilon_{kj}, \epsilon_{k'j}] = 
  \begin{cases}
    0     & \text{if } k \ne k' \\
    \sigma^2_r  & \text{if } k = k'
  \end{cases} 
\]
In this way, the within--resample correlation is estimated to be $\rho = \sigma^2/(\sigma^2+\sigma^2_r)$. This model can be fit via generalized least squares \citep{Vones:Chinc:97} to estimate the effects of the tuning parameters via the $\tau_j$ and the two variance parameters. 

The model in Equation \ref{E:delta} contains $p_k-1$ slope parameters $\tau_j$. The ``reference cell'' in this model is the current best condition at the $k^{th}$ iteration and, using this parameterization, the $\tau_j$ estimate the loss of performance for parameter setting $\theta_j$ from the current numerically optimal condition. One--sided $(1 - \alpha)$$\%$ confidence intervals can be constructed for the $\tau_j$. If the interval contains zero, this would be equivalent to failing to reject $H_0: \tau_j = 0$ versus $H_1: \tau_j > 0$. Rejection of this null indicates that the average performance of model $j$ is statistically worse than the current best model. At each iteration of resampling, any tuning parameter setting whose interval does not contain zero is removed and is not evaluated on subsequent iterations. Like Shen's model, the confidence level controls the rate at which sub--models are discarded. 

Returning to the mutagenicity example, a first evaluation of the SVM sub--models occurred after $i = 10$ splits. At the this point, the SVM model with the largest mean area under the ROC curve was $\theta = 2^{ 2 }$. The performance profile after 10 resamples was very similar to the one shown in Figure \ref{F:qsar_svm_cv}. Fixing this sub--model as the reference, $\Delta_{ij}$ values were computed for ($i \leq 10$) and these values were used in Equation \ref{E:delta}. From this model, $\widehat{\rho} = 0.34$ and $\widehat{\sigma} = 0.0043$. Using $\alpha = 0.01$, there were 6 sub--models whose intervals included zero: $2^{ 0 }$, $2^{ 0.5 }$, $2^{ 1 }$, $2^{ 1.5 }$, $2^{ 2.5 }$, $2^{ 3 }$. Therefore, the next iteration of resampling would only evaluate 7 sub--models. After this evaluation, several more models were removed: at $i = 10$, 15 models were removed, at $i = 11$, 2 models were removed and a single models were removed at $i = 13$ and $i = 14$. There were 2 surviving values of $\theta$ at $B = 50$: $2^{ 1.5 }$ and $2^{ 2 }$. The usual pick--the--winner strategy was used here to select $\theta = 2^{ 1.5 }$. 

The potential advantage of adaptive sampling can quantified by the {\em speed--up}, calculated as the total execution time for tuning the model with the complete set of resamples divided by the execution time of the adaptive procedure. For example, a speed--up of 1.5 is a fifty percent decrease in the execution time when using the adaptive procedure. For this approach and these data, a speed--up of 3.5 was achieved. This adaptive process fit 299 SVM models or 28.5$\%$ of the number required for the full set of resamples and resulted in the same choice of the SVM cost parameter.

Note that, for some models, there can be a multiple tuning parameters and this can lead to a large number of distinct combinations. The data used to conduct the futility analysis is driven by the number of model parameters ($p_i$) and the current number of resamples ($i$). It is possible that, when $p_i$ is large and $i$ is small, the generalized linear model will be over--determined and/or inestimable. Also, the assumption of normality of the residuals may be unrealistic since a highly accurate machine learning model might generate resampled performance estimates that are skewed. For example, as the area under the ROC curve approaches unity, its resampling distribution can become significantly left--skewed. Since the generalized linear model is estimating multiple variance parameters, non--normal residuals can have a profoundly adverse affect on those estimates.

\subsection{Bradley--Terry Models to Estimate Futility}\label{S:BT}

The approach shown here to measure futility is based on \citet{Consensus}, who developed models to create consensus rankings of different models based on resampled performance statistics. We modify their method to characterize the differences of tuning parameters {\em within a model}. More recently, \citet{Eugster:2013hl} used similar methods to characterize differences between models and across different data sets. 

For our purposes, the resampling data generated during the tuning process can be decomposed into a set of win/loss/tie comparisons based on the resampled performance estimates. To compare settings $\theta_j$ and $\theta_{j'}$ at resampling iteration $i$, the number of wins for  $\theta_j$ is the sum of the resamples where $Q_{ij} > Q_{ij'}$. The converse is also true for the number of wins for $\theta_{j'}$. Ties are handled as a half--win for each team. Given a set of pair--wise win/loss/tie statistics, the Bradley--Terry model \citep{bradley1952rank} is a logistic regression model where the outcome is
\[
logit[Pr(Q_{ij} > Q_{ij'})] = \lambda_j - \lambda_{j'}.
\]
The $\lambda$ parameters are estimated via maximum likelihood estimation in the usual manner. 

In the context of evaluating tuning parameter combinations, the estimated contrasts $\widehat{\lambda}_j - \widehat{\lambda}_{j'}$ can be interpreted as the log--odds that tuning parameter $\theta_j$ has a better {\em ability} to win compared to the reference setting of $\theta_{j'}$. Our approach is to use the parameter associated with the best average fitness value as the reference setting. The consequence of this choice is that most of the differences $\widehat{\lambda}_j - \widehat{\lambda}_{j'}$ will be negative and larger values indicate performance that is closer to the current best setting. It may be possible that one or more sub--models have no wins against any other sub--model. The consequence of this situation is that the ability estimates become degenerate and their associated standard errors can be orders of magnitude larger than is reasonable. To avoid this, these cases are removed from the data prior to fitting the logistic regression model and the corresponding sub--models are not considered in the remaining iterations of resampling. 

To use this approach for comparing tuning parameters, the win/loss/tie data can be used to fit the Bradley--Terry model. Similar to the approach in the previous section, one--sided (asymptotic) confidence bounds for the ability values can be calculated and used to winnow values in $\Theta$. These intervals are asymptotic and use a normal quantile $\Phi^{-1}(1-\alpha)$. The intervals produced by generalized least squares are not asymptotic and use a similar quantile of a $t$--distribution. Analogous to the linear model approach, any tuning parameter settings whose upper bound is not greater than zero would be eliminated from further consideration. This process is repeated at each resampling iteration after $B_{min}$  until either a single parameter setting remains or the maximum number of resamples is reached. 

\begin{figure}[t]
  \begin{center}  
\begin{knitrout}
\definecolor{shadecolor}{rgb}{0.969, 0.969, 0.969}\color{fgcolor}
\includegraphics[width=.8\textwidth]{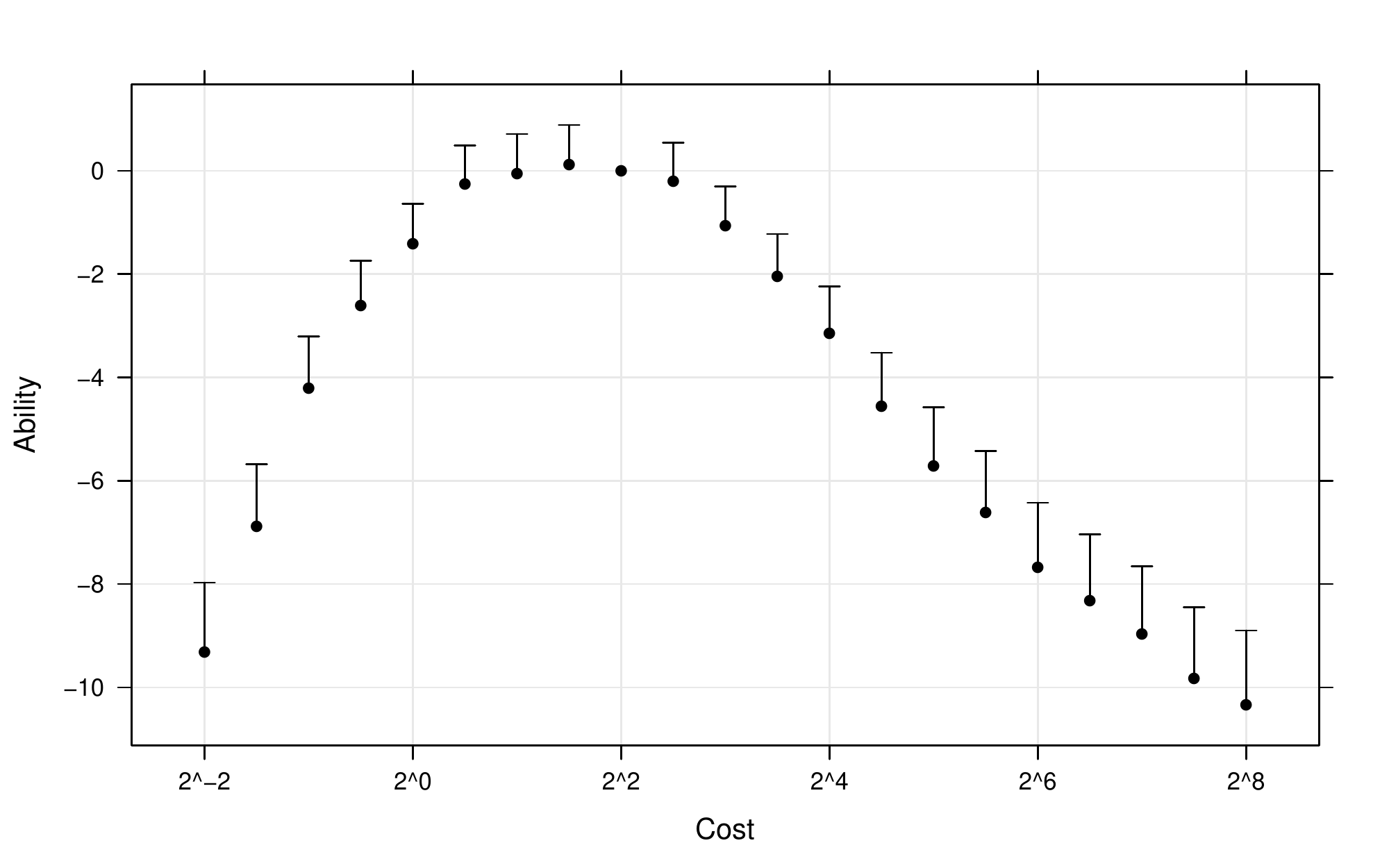} 

\end{knitrout}

    \caption{Ability estimates ($\widehat{\lambda}_j - \widehat{\lambda}_{j'}$) from the Bradley--Terry model after $B_{min} = 10$ resamples. $\theta = 2^{ 2 }$ was used as the reference model and the error bars correspond to asymptotic 95$\%$ confidence intervals. }
    \label{F:qsar_bt}
  \end{center}
\end{figure}

For the previously trained SVM model, suppose the first futility analysis was also conducted at $B_{min} = 10$. From these data, $ \binom{21}{2} = 210$ sets of ten ``tournaments'' were played between pairs of competing sub--models. The area under the ROC curve is used to compute win/loss/tie scores. For example, the model $\theta=2^{2}$ has 8 wins and 2 losses against $\theta=2^{3}$. Again, $\theta = 2^{ 2 }$ was designated as the reference model, the Bradley--Terry model was computed. Figure \ref{F:qsar_bt} shows the estimates of the ability scores for each setting along with their corresponding 95$\%$ one--sided intervals. Note that the model with the best average area under the ROC curve did not have the highest ability, illustrating the difference in average fitness values and dichotomized ``competitions'' between models. Including the reference sub--model, the conditions that survived the filtering process were $\theta = \{2^{ 0.5 }, 2^{ 1 }, 2^{ 1.5 }, 2^{ 2 }, 2^{ 2.5 }\}$. Similar to the other adaptive procedure, additional models were removed in subsequent iterations: at $i = 10$, 16 models were removed and single models were eliminated at $i = 13$, $i = 24$ and $i = 34$. There were the same 2 surviving values of $\theta$ at $B = 50$: $2^{ 1.5 }$ and $2^{ 2 }$. The same sub--model was selected as the previous two analyses. When the futility was estimated by means of the Bradley--Terry model, a speed--up of 3.2 was achieved. As before, this was largely due to fitting only 331 SVM models  (31.5$\%$ of the full set of resamples).

Since the resampled performance measures are coerced to binary values, this  may have the effect of desensitizing the adaptive procedure and result in a potential loss of inferential power. As previously noted, in situations where the fitness values are close to their boundaries (e.g. an area under the ROC curve near unity or RMSE near zero), the distribution of the $Q_{ij}$ can become significantly skewed and the linear model's assumption of normality of the residuals may not hold.  However, the dichotomization process does protect against skewness or influence of aberrantly extreme values of $Q_{ij}$. Also, the effective sample size used for the Bradley--Terry model is based on the number of pair--wise competitions. At iteration $i$ of resampling, the number of win/loss/tie statistics is $i\times\binom{p_i}{2}$. When $p_i$ is large relative to $i$, the Bradley--Terry model may be estimable where the generalized linear model may not.

\section{Simulation Studies}

To understand the potential benefits and pitfalls of these approaches, simulation studies were used. \citet{Friedman:1991p109} described a system for simulating nonlinear regression models. Four independent uniform random variables $X_1 \ldots X_4$ were used with the following regression structure:
\[
Y = atan(((X_2 X_3 - (1/(X_2 X_4)))/X_1) + \epsilon
\]
where $\epsilon \sim N(0, 0.1)$. To make the simulation more realistic, 46 independent, non--informative random normal predictors were added to the training and test sets. This system was used to evaluate the efficacy of the proposed algorithm.

For each simulated data set, an artificial neural network model \citep{Bishop95} was used to model and predict the data. A single layer feed--forward architecture was used and candidate sub--models included between 1 and 10 hidden units. Also, the model was tuned over three amounts of weight decay: 0, 10$^{-3}$ and 10$^{-2}$. The full set of 30 sub--models was evaluated for each neural network model. 

Repeated 10--fold cross--validation was used as the resampling method. The total number of resamples varied between 20 and 100 in the simulations along with:
\begin{itemize}
\item The training set size was varied: 200, 400 and 600.
\item Two settings for the minimum number of resamples ($B_{min}$) were evaluated: 10 and 20.
\item The confidence values for the confidence intervals were evaluated over three values for both adaptive procedures: $\alpha \in \{0.001, 0.01, 0.1\}$. 
\end{itemize}
For each simulation setting, a minimum 100 data sets were created and analyzed (the final number was affected by hardware failures). The simulated test set ($n$ = 100,000) root mean squared error was calculated for models corresponding to the complete set of resamples and adaptive strategies. The efficacy of the two adaptive procedures were quantified by the speed--up and the difference in RMSE between the full set of resamples and the adaptive procedures. The computations were conducted in \pkg{R} using a modified version of the \pkg{caret} package \citep{caret}. 

\begin{figure}[t]
  \begin{center}  
\begin{knitrout}
\definecolor{shadecolor}{rgb}{0.969, 0.969, 0.969}\color{fgcolor}
\includegraphics[width=.8\textwidth]{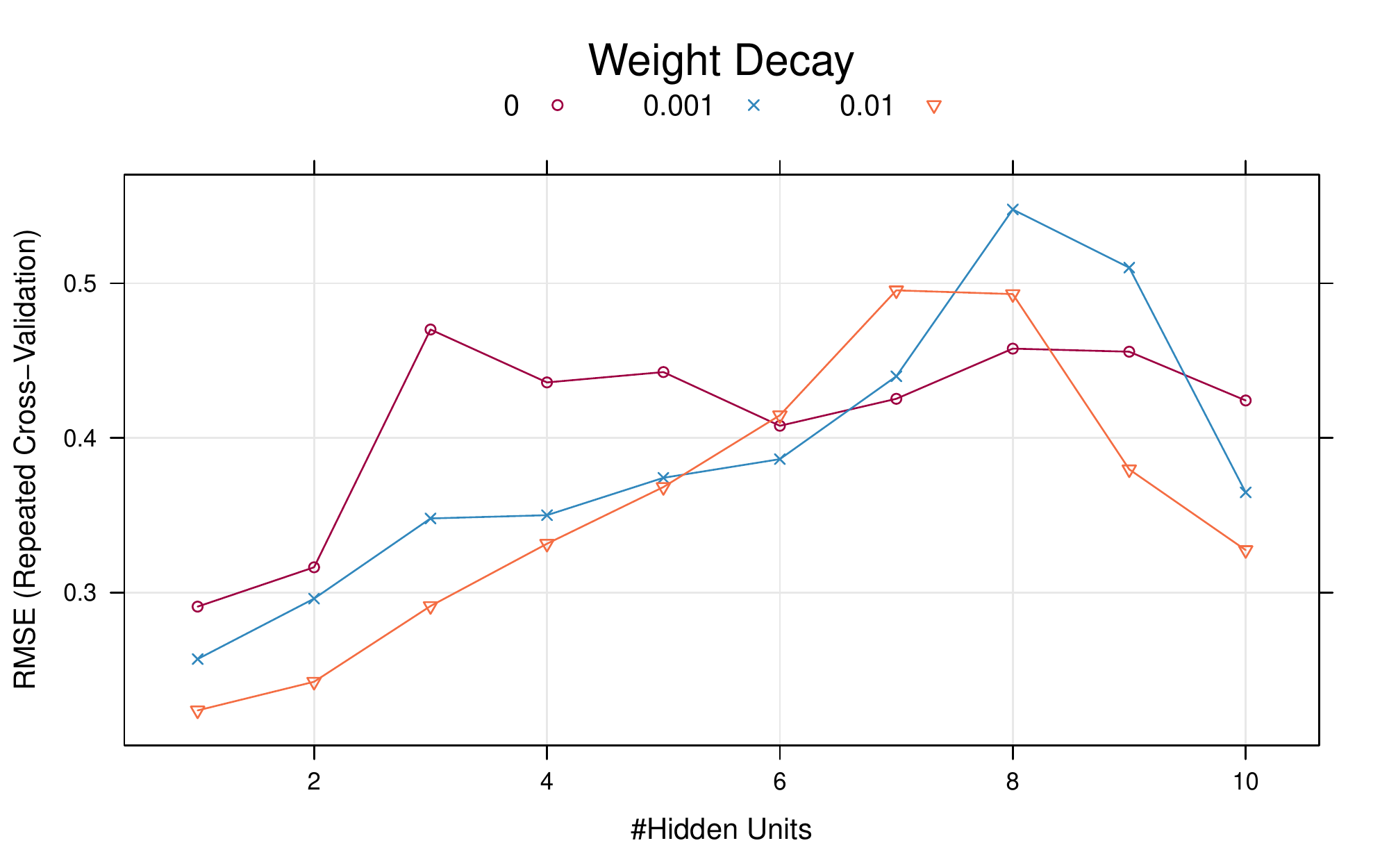} 

\end{knitrout}

    \caption{An example of a simulated resampling profile for a neural network model using $B=100$ resamples generated with repeated 10--fold cross--validation.}
    \label{F:sim_nnet_profile}
  \end{center}
\end{figure}

To illustrate the relationship between performance and the tuning parameters, one simulated model with 1000 training set points was tuned with six repeats of 10--fold cross--validation. The resulting profile can be seem in Figure \ref{F:sim_nnet_profile}. Based on these data, the optimal parameter settings are a single hidden unit and a weight decay value of 0.01. The other settings that are most competitive with this condition have one or two hidden units and many of the other settings are likely to be eliminated quickly, depending on how much uncertainty is associated with the $Q_{kj}$.

\begin{figure}
  \begin{center}  
\begin{knitrout}
\definecolor{shadecolor}{rgb}{0.969, 0.969, 0.969}\color{fgcolor}
\includegraphics[width=.9\textwidth]{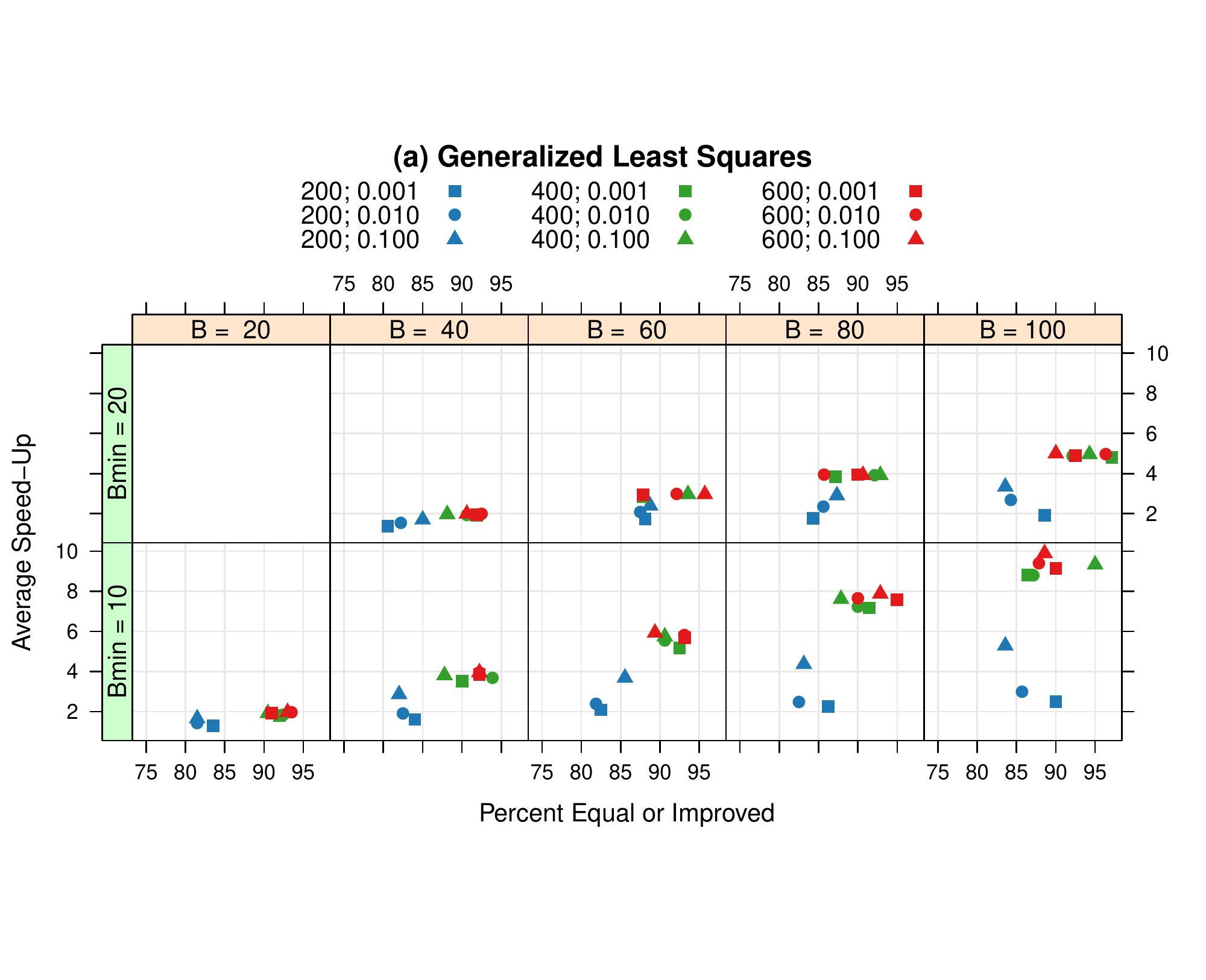} 

\end{knitrout}

\vspace{-1.1in}

\begin{knitrout}
\definecolor{shadecolor}{rgb}{0.969, 0.969, 0.969}\color{fgcolor}
\includegraphics[width=.9\textwidth]{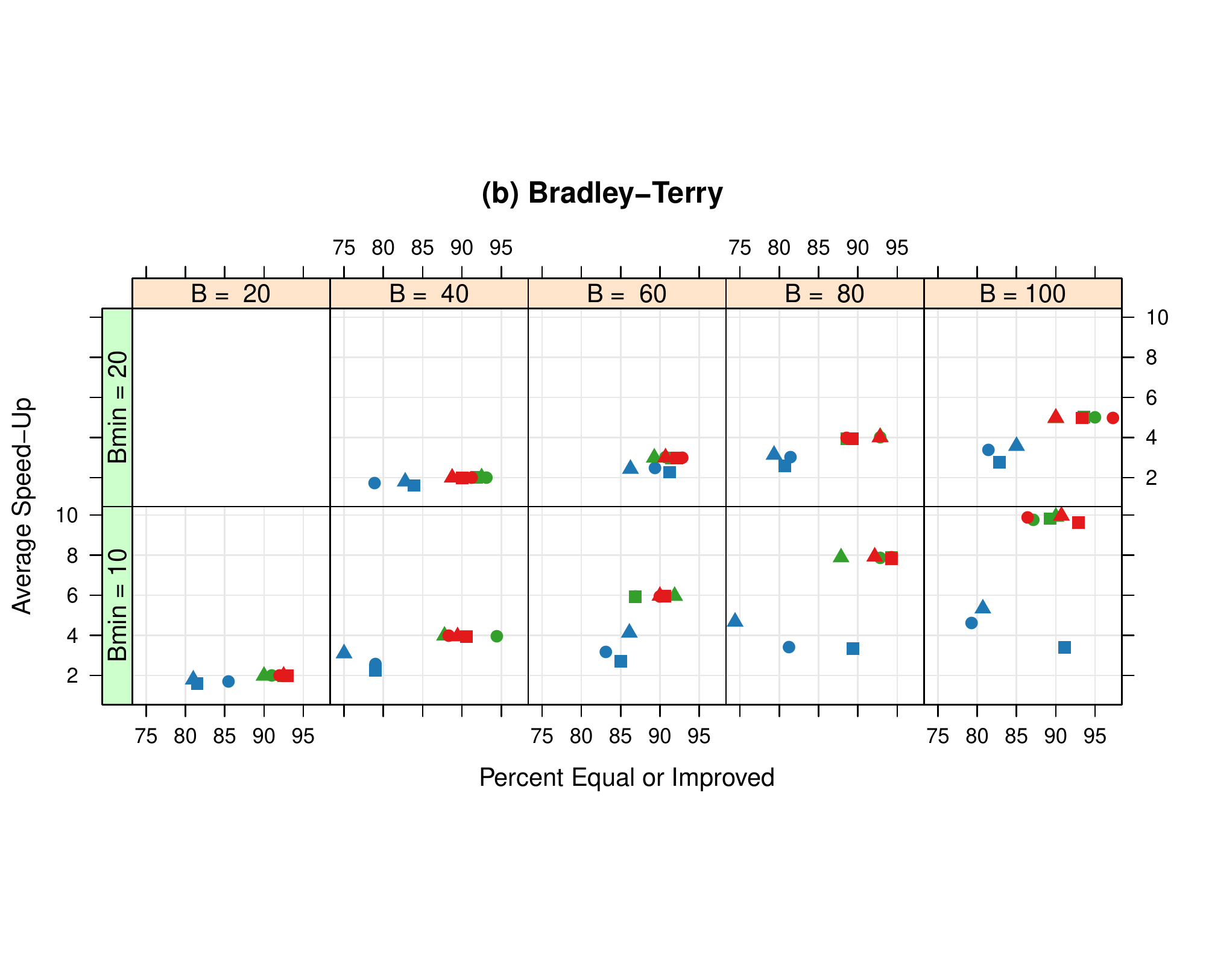} 

\end{knitrout}

\vspace{-.5in}

    \caption{The median speed--up versus the percentage of simulations where the adaptive procedure selected a model at least as good as the full resampling process. The legend indicates the training set size (either 200, 400 or 600) and the value of $\alpha$ (either 0.001, 0.01 or 0.1).}
    \label{F:speedup_seq}
  \end{center}
\end{figure}

In general, the concordance between the two adaptive procedures and the nominal approach with the full set of resamples was  good. Using linear models, 81.9$\%$ of the final parameters matched the settings found with the nominal procedure (across all simulated data sets and conditions). Similarly, 82$\%$ of the model settings match found via the Bradley--Terry model matched. However, the full resampling process is not infallible may not yield the best test set results. Figure \ref{F:speedup_seq} shows the percentages of simulations where the adaptive procedures selected the same sub--model as the nominal resampling scheme {\it or} the adaptive resampling chose a model with {\em better} test set results. Based on this revised criterion, the overall percentage of models at least as good as the matching fully resampled model increased to 88.9$\%$ and 88.2$\%$ for the linear models and Bradley--Terry approach, respectively. 

From the plots in Figure \ref{F:speedup_seq}, there are several patterns. First, the training set size has the most significant affect on the probability of a good model. The smallest training set size ($n=200$) has the worst efficacy and the two larger sizes had roughly comparable findings in terms of choosing a good model. This is most likely due to the quality of the estimated values of $Q_{ij}$ since larger training set sizes lead to larger holdout sets. As the accuracy in the $Q_{ij}$ increases, the likelihood of discarding a quality value of $\theta_j$ decreases. In these simulations, $B_{min}$ and the confidence level did not appear to have a major impact on the quality of the model within the ranges that were studied here. For values of $B > 20$, the number of resamples did not appear to have much of an effect. Finally, the method of computing futility showed comparable value.  

The speed--up of the procedures are also shown in these figures. The median speed-up over all simulations were 2.9 and 3.1 for the generalized least squares and Bradley--Terry methods, respectively.
Similarly, the best case speed--up, where the training set size and the number of resamples are large, were 26.3 and  33.1, respectively. There were a small number of simulations (0.5\% of the total) that took longer with the adaptive procedures and were more likely to occur using generalized linear models. This occurred with small training sets and fewer resamples. Overall, the speed--ups were driven by the number of resamples ($B$) and the training set size. This makes intuitive sense as these two factors are surrogates for the total computational cost of model tuning.

\section{The Effect of Parallel Processing}

Parallel processing has become more common in scientific computing. Many computers are currently configured with multicore architectures and open--source software is widely available to run computations in parallel \citep{Schmidberger:2009tx,mccallum2011parallel}. The model tuning process is ``embarrassingly parallel''. In Algorithm \ref{A:Resamp}, the two {\tt for} loops (lines 2 and 4) are not serial, meaning the computations inside of the loops are independent. For the SVM model described in Section \ref{S:intro}, a total of 1050 models were fit across different sub--model configurations and resamples. There is no logical barrier to running these computations in parallel and doing so leads to substantial speed--up \citep{apm}. For the non--adaptive approach, the result of parallel processing with six {\em worker} processes was a speed--up of 3--fold (relative to sequentially processing the full set of resamples). 
Reasonable questions would be ``can I get the same time reduction using parallel processing and the full set of resamples? or ``does the adaptive procedure still offer advantages in parallel?''

The process shown in Algorithm \ref{A:adaptive} can also benefit from parallel processing. First, all of the computations across models (i.e. those within lines 4 and 6 of Algorithm \ref{A:adaptive}) can be conducted in parallel. The only situation where the resamples cannot be run in parallel are when $i > B_{min}$ and  $p_i > 1$. However, there is some slow--down associated with the additional computations required to conduct the futility analysis. 
 
For the SVM model, the adaptive techniques were also evaluated with six worker processes. When compared to the full resampling approach run in parallel, the speed--up for the linear model and Bradley--Terry approaches were 3.6 and 3.5, respectively. This indicates that there is  benefit to the adaptive procedures above and beyond those imparted using parallel processing. 
 
The simulation studies were repeated with parallel processing. In this study, multicore forking of calculations \citep{Schmidberger:2009tx,Eugster:tr} is used to run combinations of models and resampled data sets using more than one processor on the same machine. Version 0.1.7 of \pkg{R}'s \pkg{multicore} package was used. In these simulations, the computational tasks were split over six sub--processes as previously described. 

\begin{figure}
  \begin{center}  
\begin{knitrout}
\definecolor{shadecolor}{rgb}{0.969, 0.969, 0.969}\color{fgcolor}
\includegraphics[width=.9\textwidth]{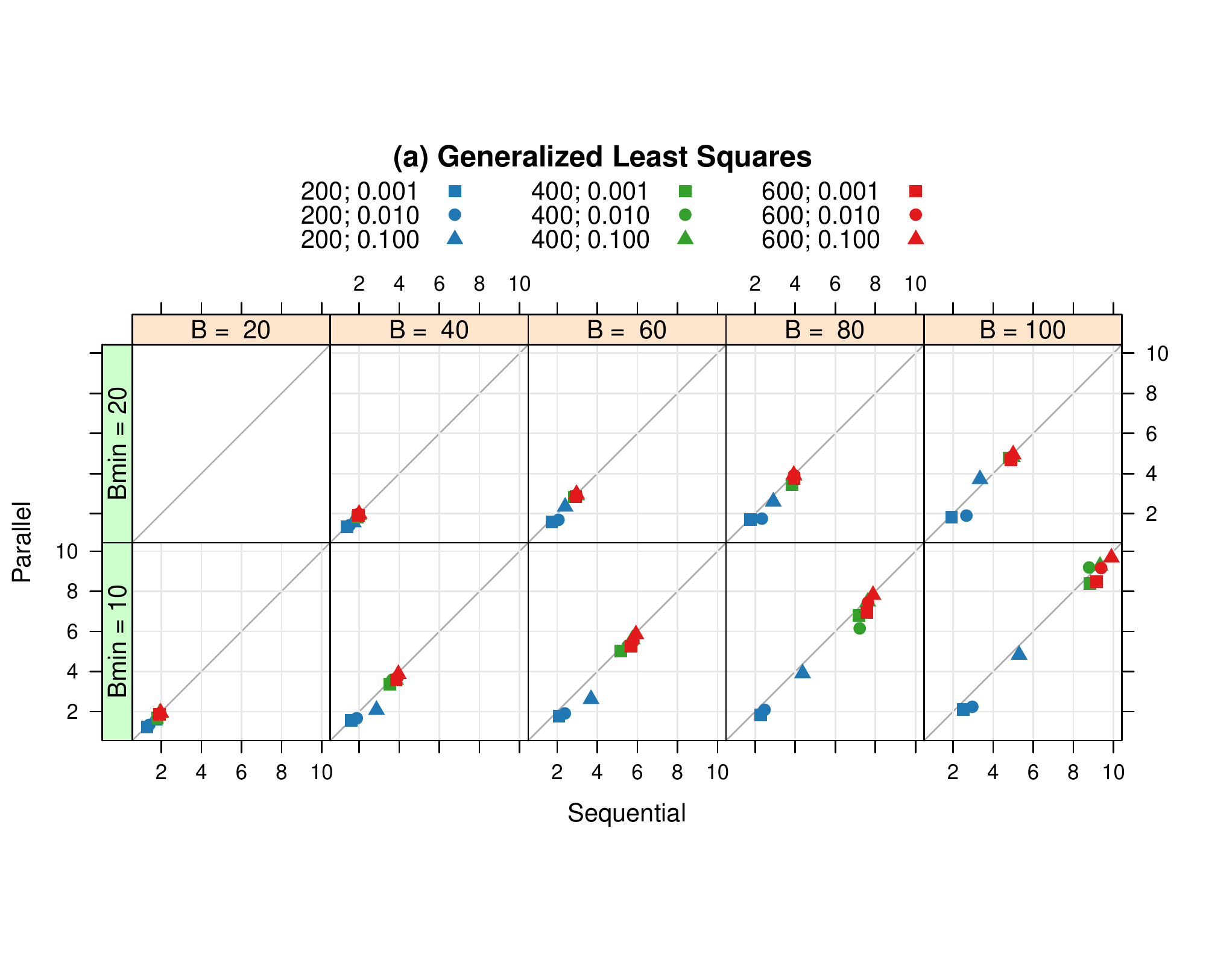} 

\end{knitrout}

\vspace{-1.2in}

\begin{knitrout}
\definecolor{shadecolor}{rgb}{0.969, 0.969, 0.969}\color{fgcolor}
\includegraphics[width=.9\textwidth]{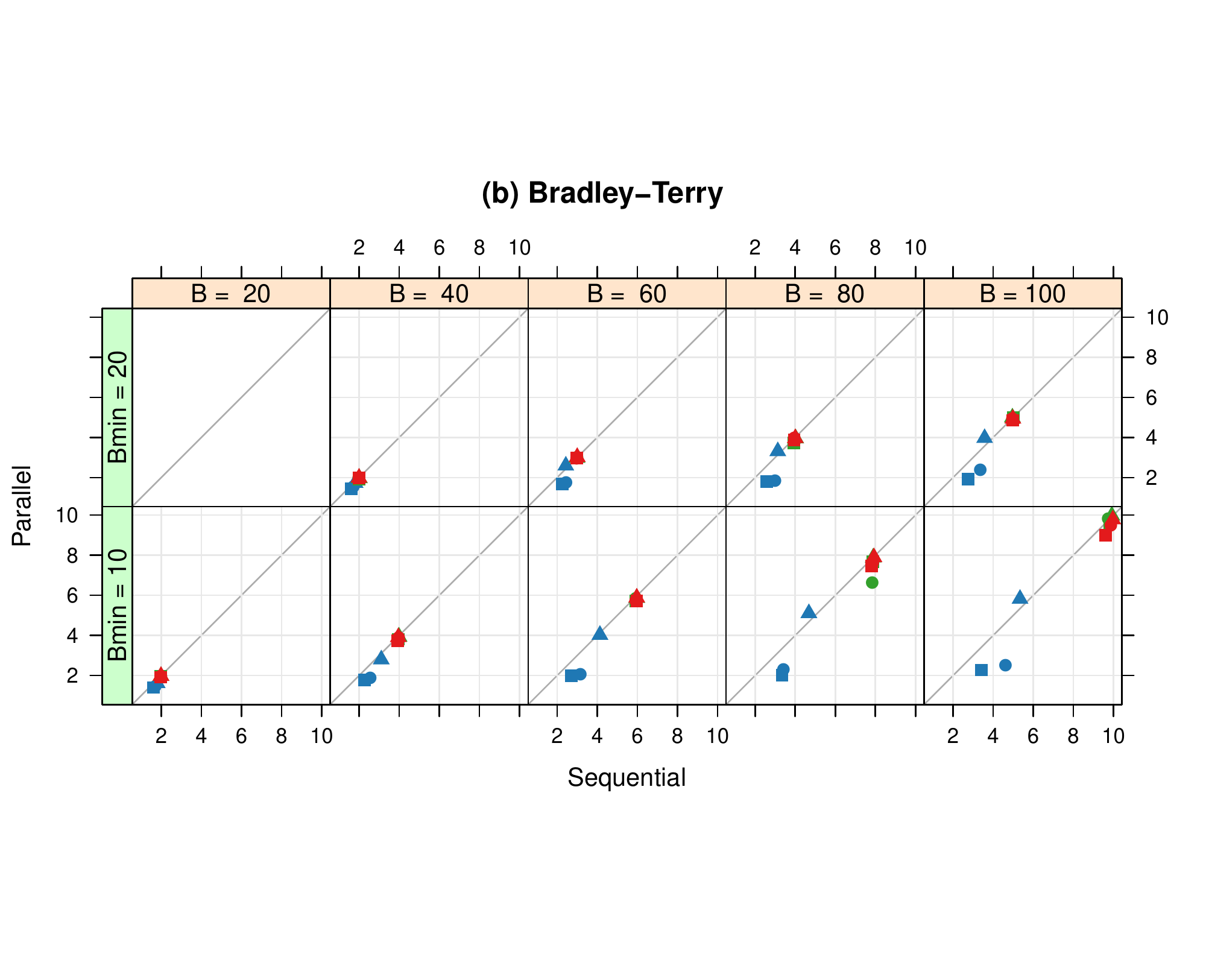} 

\end{knitrout}

\vspace{-.5in}

    \caption{The median speed--ups for the sequential and parallel computations.}
    \label{F:speedup_par}
  \end{center}
\end{figure}

Figure \ref{F:speedup_par} shows the average speed--ups for the sequential and parallel computations. Under these conditions, the median speed-up over all simulations were 3.2 and 3.5 for the generalized least squares and Bradley--Terry methods, respectively. For both adaptive procedures, there was a high degree of correlation in the median speed--ups. This indicate that, for this simulation, parallel processing did not eliminate the benefits of adaptively removing tuning parameter values.

Since the speed--ups are comparable between different technologies, this indicates that,  independent of the technology,  adaptive methods are faster. For example, in the generalized least squares simulations with 600 points in the training set, $B = 60$, $B_{min} = 20$ and $\alpha = 0.01$, the median time to get the full set of resamples sequentially was 40.9 hours. Using parallel processing only, the median time was reduced to 11.3 hours. Without parallel processing, adaptive resampling would have reduced the training time to 13.8 hours. However, the biggest savings occurred with adaptive resampling in parallel; here the median time was 3.9 hours.

\section{Discussion}

In this manuscript, a resampling scheme was described that is effective at finding reasonable values of tuning parameters in a more computationally efficient manner. The efficacy and efficiencies of the procedures are best when the training set size is not small and computational cost of fitting the sub--models is moderate to high. The computational gains afforded by parallel processing technologies do not obviate the gains in the proposed methodologies. 

There are several possible improvements to the adaptive procedures that were not explored here. For some models, there is the possibility that several values of $\theta$ will produce identical models. For example, some decision trees are tuned over their {\em maximum} possible depth but, after pruning, the same tree may result. As another example, multivariate adaptive regression spline \citep{Friedman:1991p109} models conduct feature selection to remove model terms. Because of this, it is possible for multiple values in $\Theta$ to generate the same model predictions and thus have identical fitness values. The consequence of this is that the futility analysis procedure will never be able to differentiate between these models and may conduct unnecessary iterations of resampling. To mitigate these risks, a one--time filter could be used to remove values of $\theta$ that are identical (within floating point precision) when $i \ge B_{min}$. 

Also, if the modeling goal was strictly {\it cross--validatory choice} and the adaptive procedure determines eliminates all but a single setting, there may be no need to estimate the performance for the model using all $B$ resamples. In such cases, the speed--up values would substantially increase. 

Finally, if there is a string desire to minimize the possibility that the adaptive procedure might choose a sub--optimal model, a more conservative approach could be used. For example, this manuscript used a static confidence level. Alternatively, a dynamic approach where the likelihood of discarding settings is small at the beginning of the adaptive procedure, but increases at each iteration, may be more appropriate. 

\vskip 0.2in

\bibliographystyle{plainnat}
\bibliography{adaptive}

\begin{thebibliography}{33}
\providecommand{\natexlab}[1]{#1}
\providecommand{\url}[1]{\texttt{#1}}
\expandafter\ifx\csname urlstyle\endcsname\relax
  \providecommand{\doi}[1]{doi: #1}\else
  \providecommand{\doi}{doi: \begingroup \urlstyle{rm}\Url}\fi

\bibitem[Altman and Bland(1994)]{Altman:1994uv}
D~Altman and J~Bland.
\newblock {Diagnostic tests 3: Receiver Operating Characteristic plots.}
\newblock \emph{British Medical Journal}, 309\penalty0 (6948):\penalty0 188,
  1994.

\bibitem[Bishop(1995)]{Bishop95}
C~Bishop.
\newblock \emph{Neural Networks for Pattern Recognition}.
\newblock Clarendon Press, Oxford, 1995.

\bibitem[Bishop(2007)]{Bishop2007}
C~Bishop.
\newblock \emph{{Pattern Recognition and Machine Learning}}.
\newblock Springer, 2007.

\bibitem[Bradley and Terry(1952)]{bradley1952rank}
R~Bradley and M~Terry.
\newblock Rank analysis of incomplete block designs: I. the method of paired
  comparisons.
\newblock \emph{Biometrika}, 39\penalty0 (3/4):\penalty0 324--345, 1952.

\bibitem[Breiman(2001)]{Breiman:2001wl}
L~Breiman.
\newblock {Statistical modeling: The two cultures}.
\newblock \emph{Statistical Science}, 16\penalty0 (3):\penalty0 199--215, 2001.

\bibitem[Breiman et~al.(1984)Breiman, Friedman, Olshen, and Stone]{cart}
L~Breiman, J.~Friedman, R~Olshen, and C~Stone.
\newblock \emph{{Classification and Regression Trees}}.
\newblock Chapman and Hall, New York, 1984.

\bibitem[Brown and Davis(2006)]{Brown:2006wp}
C~Brown and H~Davis.
\newblock {Receiver Operating Characteristics curves and related decision
  measures: A tutorial}.
\newblock \emph{Chemometrics and Intelligent Laboratory Systems}, 80\penalty0
  (1):\penalty0 24--38, 2006.

\bibitem[Caputo et~al.(2002)Caputo, Sim, Furesjo, and Smola]{Caputo}
B~Caputo, K~Sim, F~Furesjo, and A~Smola.
\newblock Appearance--based object recognition using {SVMs}: Which kernel
  should {I} use?
\newblock In \emph{Proceedings of {NIPS} Workshop on Statitsical Methods for
  Computational Experiments in Visual Processing and Computer Vision}, 2002.

\bibitem[Efron(1983)]{Efron:1983ul}
B~Efron.
\newblock {Estimating the error rate of a prediction rule: Improvement on
  cross-validation}.
\newblock \emph{Journal of the American Statistical Association}, pages
  316--331, 1983.

\bibitem[Eugster et~al.(2011)Eugster, Knaus, Porzelius, Schmidberger, and
  Vicedo]{Eugster:tr}
M~Eugster, J~Knaus, C~Porzelius, M~Schmidberger, and E~Vicedo.
\newblock Hands-on tutorial for parallel computing with \pkg{R}.
\newblock \emph{Computational Statistics}, 26\penalty0 (2):\penalty0 219--239,
  2011.

\bibitem[Eugster et~al.(2013)Eugster, Leisch, and Strobl]{Eugster:2013hl}
M~Eugster, F~Leisch, and C~Strobl.
\newblock {(Psycho-)analysis of benchmark experiments: A formal framework for
  investigating the relationship between data sets and learning algorithms}.
\newblock \emph{Computational Statistics and Data Analysis}, August 2013.

\bibitem[Fawcett(2006)]{Fawcett:2006gr}
T~Fawcett.
\newblock {An introduction to ROC analysis}.
\newblock \emph{Pattern Recognition Letters}, 27\penalty0 (8):\penalty0
  861--874, 2006.

\bibitem[Fernandez-Delgado et~al.(2014)Fernandez-Delgado, Cernadas, and
  Barro]{Delgado14}
M~Fernandez-Delgado, E~Cernadas, and S~Barro.
\newblock Do we need hundreds of classifiers to solve real world classification
  problems?
\newblock \emph{Journal of Machine Learning Research (accepted; in press)},
  2014.

\bibitem[Friedman(1991)]{Friedman:1991p109}
J~Friedman.
\newblock Multivariate adaptive regression splines.
\newblock \emph{The Annals of Statistics}, 19\penalty0 (1):\penalty0 1--141,
  1991.

\bibitem[Hornik and Meyer(2007)]{Consensus}
K~Hornik and D~Meyer.
\newblock Deriving consensus rankings from benchmarking experiments.
\newblock In R~Decker and H-J. Lenz, editors, \emph{Advances in Data Analysis},
  Studies in Classification, Data Analysis, and Knowledge Organization, pages
  163--170. Springer Berlin Heidelberg, 2007.

\bibitem[Kazius et~al.(2005)Kazius, McGuire, and Bursi]{Kazius:2005up}
J~Kazius, R~McGuire, and R~Bursi.
\newblock {Derivation and validation of toxicophores for mutagenicity
  prediction}.
\newblock \emph{Journal of Medicinal Chemistry}, 48\penalty0 (1):\penalty0
  312--320, 2005.

\bibitem[Kim(2009)]{Kim:2009im}
J-H Kim.
\newblock {Estimating classification error rate: Repeated cross-validation,
  repeated hold-out and bootstrap}.
\newblock \emph{Computational Statistics and Data Analysis}, 53\penalty0
  (11):\penalty0 3735--3745, September 2009.

\bibitem[Kohavi and John(1997)]{Kohavi:1997wca}
R~Kohavi and G~John.
\newblock {Wrappers for feature subset selection}.
\newblock \emph{Artificial Intelligence}, 97\penalty0 (1):\penalty0 273--324,
  1997.

\bibitem[Krstajic et~al.(2014)Krstajic, Buturovic, Leahy, and
  Thomas]{krstajic2014cross}
D~Krstajic, L~Buturovic, D~Leahy, and S~Thomas.
\newblock Cross--validation pitfalls when selecting and assessing regression
  and classification models.
\newblock \emph{Journal of cheminformatics}, 6\penalty0 (1):\penalty0 1--15,
  2014.

\bibitem[Kuhn(2008)]{caret}
M~Kuhn.
\newblock Building predictive models in \pkg{R} using the \pkg{caret} package.
\newblock \emph{Journal of Statistical Software}, 28\penalty0 (5):\penalty0
  1--26, 2008.

\bibitem[Kuhn and Johnson(2013)]{apm}
M~Kuhn and K~Johnson.
\newblock \emph{{Applied Predictive Modeling}}.
\newblock Springer, 2013.

\bibitem[Lachin(2005)]{Lachin:2005uu}
J~Lachin.
\newblock {A review of methods for futility stopping based on conditional
  power}.
\newblock \emph{Statistics in Medicine}, 24\penalty0 (18):\penalty0 2747--2764,
  2005.

\bibitem[Leach and Gillet(2007)]{Leach:2007tq}
A~Leach and V~Gillet.
\newblock \emph{{An Introduction to Chemoinformatics}}.
\newblock Springer, 2007.

\bibitem[McCallum and Weston(2011)]{mccallum2011parallel}
E~McCallum and S~Weston.
\newblock \emph{Parallel \pkg{R}}.
\newblock O'Reilly Media, Inc., 2011.

\bibitem[Molinaro(2005)]{Molinaro:2005p47}
A~Molinaro.
\newblock Prediction error estimation: A comparison of resampling methods.
\newblock \emph{Bioinformatics}, 21\penalty0 (15):\penalty0 3301--3307, 2005.

\bibitem[Quinlan(1993)]{quinlan1993c4}
R~Quinlan.
\newblock \emph{\textsf{C4.5}: Programs for Machine Learning}.
\newblock Morgan Kaufmann Publishers, 1993.

\bibitem[Schmidberger et~al.(2009)Schmidberger, Morgan, Eddelbuettel, Yu,
  Tierney, and Mansmann]{Schmidberger:2009tx}
M~Schmidberger, M~Morgan, D~Eddelbuettel, H~Yu, L~Tierney, and U~Mansmann.
\newblock State--of--the--art in parallel computing with \pkg{R}.
\newblock \emph{Journal of Statistical Software}, 31\penalty0 (1):\penalty0
  1--27, 2009.

\bibitem[Shen et~al.(2011)Shen, Welch, and Hughes-Oliver]{Shen:2011vi}
H~Shen, W~Welch, and J~Hughes-Oliver.
\newblock {Efficient, adaptive cross--validation for tuning and comparing
  models, with application to drug discovery}.
\newblock \emph{The Annals of Applied Statistics}, 5\penalty0 (4):\penalty0
  2668--2687, 2011.

\bibitem[Snapinn et~al.(2006)Snapinn, Chen, Jiang, and
  Koutsoukos]{Snapinn:2006ih}
S~Snapinn, M-G Chen, Q~Jiang, and T~Koutsoukos.
\newblock {Assessment of futility in clinical trials}.
\newblock \emph{Pharmaceutical Statistics}, 5\penalty0 (4):\penalty0 273--281,
  2006.

\bibitem[{\"U}st{\"u}n et~al.(2005){\"U}st{\"u}n, Melssen, Oudenhuijzen, and
  Buydens]{Ustun:2005ew}
B~{\"U}st{\"u}n, W~J Melssen, M~Oudenhuijzen, and L~M~C Buydens.
\newblock {Determination of optimal support vector regression parameters by
  genetic algorithms and simplex optimization}.
\newblock \emph{Analytica Chimica Acta}, 544\penalty0 (1-2):\penalty0 292--305,
  July 2005.

\bibitem[Vapnik(2010)]{vapnik2010nature}
V~Vapnik.
\newblock \emph{The Nature of Statistical Learning Theory}.
\newblock Springer, 2010.

\bibitem[Vonesh and Chinchilli(1997)]{Vones:Chinc:97}
E~Vonesh and V~Chinchilli.
\newblock \emph{Linear and Nonlinear Models for the Analysis of Repeated
  Measurements}.
\newblock Marcel Dekker, New York, 1997.

\bibitem[Wold et~al.(2001)Wold, Sostrom, and Eriksson]{wold2001pls}
S~Wold, M~Sostrom, and L~Eriksson.
\newblock {PLS}--regression: a basic tool of chemometrics.
\newblock \emph{Chemometrics and Intelligent Laboratory Systems}, 58\penalty0
  (2):\penalty0 109--130, 2001.

\end{thebibliography}

\end{document}